\documentclass[twoside,11pt]{article}
\usepackage{jair, theapa, rawfonts}

\usepackage{amsfonts}
\usepackage{amsthm}
\usepackage{amsmath}
\usepackage{amssymb}
\usepackage{booktabs}
\usepackage{multirow}
\usepackage{xcolor}
\usepackage{soul}
\usepackage[linesnumbered,vlined,ruled]{algorithm2e}
\usepackage{tikz}
\usepackage{tkz-graph}
\SetVertexNormal[Shape      = circle,
                 FillColor  = orange,
                 LineWidth  = 2pt]
\SetUpEdge[lw         = 1.5pt,
           color      = black,
           labelcolor = white,
           labeltext  = red,
           labelstyle = {sloped,draw,text=blue}]
           
\usepackage[switch]{lineno}  
\usetikzlibrary{arrows, decorations.markings}
\newcommand\norm[1]{\left\lVert#1\right\rVert}
\tikzstyle{vecArrow} = [thick, decoration={markings,mark=at position
   1 with {\arrow[semithick]{open triangle 60}}},
   double distance=2pt, shorten >= 5.5pt,
   preaction = {decorate},
   postaction = {draw,line width=2pt, white,shorten >= 4.5pt}]
\tikzstyle{innerWhite} = [semithick, white,line width=2pt, shorten >= 4.5pt]

\newtheorem{proposition}{Proposition}
\DeclareMathOperator*{\argmax}{arg\,max}
\DeclareMathOperator*{\argmin}{arg\,min}
\makeatletter
\DeclareRobustCommand{\sqcdot}{\mathbin{\mathpalette\morphic@sqcdot\relax}}
\newcommand{\morphic@sqcdot}[2]{%
	\sbox\z@{$\m@th#1\centerdot$}%
	\ht\z@=.33333\ht\z@
	\vcenter{\box\z@}%
}

\jairheading{78}{2023}{823-848}{07/2023}{11/2023}
\ShortHeadings{Graphmax for Text Generation}
{Liu \& Yin}
\firstpageno{823}

\begin{document}

\title{Graphmax for Text Generation}

\author{\name Bin Liu \email liubin@swufe.edu.cn \\       \addr The Center of Statistical Research, School of Statistics,\\ Southwestern University of Finance and Economics, Chengdu, China
       \AND
       \name Guosheng Yin \email gyin@hku.hk  \\
       \textit{Corresponding author}\\
       \addr Department of Statistics and Actuarial Science,\\
	The University of Hong Kong, Hong Kong, China
       }

\maketitle

\begin{abstract}
In text generation, a large language model (LM) makes a choice of each new word based only on the former selection of its context using the \texttt{softmax} function. Nevertheless, the link statistics information of concurrent words based on a scene-specific corpus is valuable in choosing the next word, which can help to ensure the topic of the generated text to be aligned with the current task. To fully explore the co-occurrence information, we propose a \texttt{graphmax} function for task-specific text generation. Using the graph-based regularization, \texttt{graphmax} enables the final word choice to be determined by both the global knowledge from the LM and the local knowledge from the scene-specific corpus. The traditional \texttt{softmax} function is regularized with a graph total variation (GTV) term, which incorporates the local knowledge into the LM and encourages the model to consider the statistical relationships between words in a scene-specific corpus.
The proposed \texttt{graphmax} is versatile and can be readily plugged into any large pre-trained LM for text generation and machine translation.
Through extensive experiments, we demonstrate that the new GTV-based regularization can improve performances in various natural language processing (NLP) tasks in comparison with existing methods. Moreover, through human experiments, we observe that participants can easily distinguish the text generated by \texttt{graphmax} or \texttt{softmax}.
\end{abstract}

\section{Introduction}
\label{Introduction}
The \texttt{softmax} operator is a fundamental component of text generation models, particularly in deep neural language models \shortcite{sutskever2011generating,radford2019language}. In these models, a context embedding $\mathbf{z}$ is generated by the hidden layers of a deep neural network, such as a recurrent neural network \cite{sutskever2011generating} or a transformer \cite{radford2019language}. The output layer, which is typically a fully connected linear layer, is connected to the \texttt{softmax} function to compute a probability distribution for the next word choice.

The \texttt{softmax} is a continuous mapping function that transforms an $n$-dimensional input vector onto an $(n-1)$-simplex. The commonly adopted setting of \texttt{softmax} for text generation imposes a strong hypothesis that the probability of the next word is globally continuous over all the possible words in a dictionary. In linguistics, the arrangement of words and phrases needs to comply with syntax or popular language rules to create fluent and meaningful sentences, which should match the human habitual ways of expression. For example, when the context is a smartphone review in an e-commerce site, people prefer to using internet-style words rather than a formal style.  
In other words, the best selection of the next word could only be achieved by incorporating the human expression habits as reflected by the current corpus. We refer to the human preference of expression of word choices as local fluency information. However, the traditional \texttt{softmax} function fails to exploit such fluency knowledge. 

The total variation (TV) shows good performance in modeling the local spatial information,
which has been successfully applied to image processing tasks \shortcite{lellmann2009convex,chambolle2010introduction,jia2019regularized}. A naive way of computing TV over 2D signal images is given by 
\begin{equation*}	\textrm{TV}_{\mathbb{R}^2} = \sum_{i,j} \sqrt{|x_{i+1,j}-x_{i,j}|^2+|x_{i,j+1}-x_{i,j}|^2},
\end{equation*}
where $x_{i,j}$ is a pixel value in a 2D image, and the subindices $i,j$ are the coordinates of this pixel \cite{rudin1992nonlinear}. In image processing, it is shown to produce piecewise smoothness regularization within a bounded $\mathbb{R}^2$ space. 

\begin{figure*}[t!]
	\centering
	\begin{tikzpicture}
		\draw (9, 0.1) node[inner sep=0] (b) {\includegraphics[width=7cm]{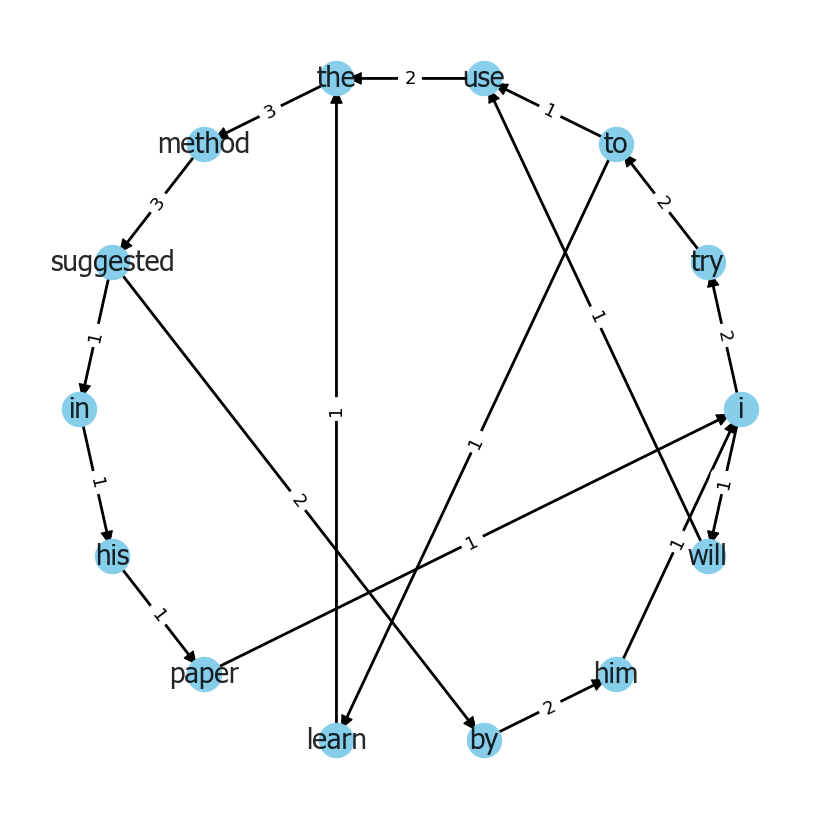}};
		\node[text width=4.4cm] at (,1.2)
		{\fontfamily{pag}\selectfont {\footnotesize I try to use the method suggested in his paper. }};
		\node[text width=4.5cm] (a) at (,0.0)
		{\fontfamily{pag}\selectfont {\footnotesize  I try to learn the method suggested by him. }};
		\node[text width=4.5cm] at (,-1.2)
		{\fontfamily{pag}\selectfont {\footnotesize I will use the method suggested by him. }};
		\node[text width=10cm] at (5,2.2){\textcolor{blue}{Corpus}};
		\draw[vecArrow] (a) to (b);
	\end{tikzpicture}
	\caption{Construction of a weighted directed graph $G = (V, \mathbf{A}, \mathcal{C})$ (right panel) from a corpus (left panel), where $V = \{v_i\}_{i=1}^N$ is a set of nodes corresponding to the words in the dictionary of the corpus, $\mathbf{A}$ is an $N\times N$ weighted adjacency matrix, and $\mathcal{C}$ is the corpus that graph $G$ is built upon. The directed edges denote the 2-grams of the corpus, where the weights are calculated based on the frequency of 2-grams in the corpus. A general $n$-gram model can be easily modeled by an $n$-order adjacency matrix $\mathbf{A}^n$.}
	\label{fig:demoGraph}
\end{figure*}

We extend the total variation to natural language processing (NLP) tasks to explore the concurrent words (or fluency) information. In particular, we define a graphical total variation for the 1D sequential text data, which can quantify the local word choice variation based on a scene-specific corpus. As shown in Figure \ref{fig:demoGraph}, the 
language rules and special expression preferences for word order choices in the corpus can be statistically quantified by a weighted directed word concurrent graph $G = (V, \mathbf{A}, \mathcal{C})$, where $V = \{v_i\}_{i=1}^N$ is a set of nodes (words), $\mathbf{A} \in \mathbb{R}^{N\times N}$ is a weighted adjacency matrix, and $\mathcal{C}$ is the corpus from which the graph is constructed.
Suppose that a mapping $\mathcal{F}:\mathcal{C} \longrightarrow G$  counts all the 2-gram tuples $(w_t,w_{t+1})$ of the sentences in $\mathcal{C}$ and then maps them to the directed edges of graph $G$ as shown in Figure \ref{fig:demoGraph}, where $w_i$ is a word in the dictionary $\mathcal{D}$, corresponding to the vertex $v_i$ of $G$.  
In practice, $\mathbf{A}_{i,j}$ is calculated by the mapping $\mathcal{F}$ based on the frequency of the corresponding 2-gram tuples $(w_t,w_{t+1})$ appeared in all the sentences of the corpus $\mathcal{C}$. Let $\mathbf{x}$ denote a graph signal over $G = (V, \mathbf{A}, \mathcal{C})$, and then we can calculate the graph-shifted text signal as 
$\mathbf{A}\mathbf{x}$ \shortcite{chen2014signal}. We define the graph total variation (GTV) for the text data as $\|\mathbf{x}-\mathbf{A}\mathbf{x}\|_2$. In text generation, the graph signal $\mathbf{x}$ can be a probability distribution for the choice of the next word.

As discussed earlier, the traditional \texttt{softmax} function aims to link the dense output of a deep neural language model (LM) onto a probability simplex. It is globally continuous but fails to incorporate the local task-specific preferences into an LM explicitly. As a remedy, we propose a GTV-regularized \texttt{softmax} function, called \texttt{graphmax}, which can be incorporated into any global pre-trained LM to improve the local fluent satisfaction of the generated sentences. For example, the \texttt{graphmax} can be integrated into GPT-2 \cite{radford2019language} and BART \shortcite{lewis2020bart} to obtain a scene-based text generation model. 
The contributions of our work are three-fold:
\begin{enumerate}
	\item We utilize GTV to capture the local style of scene-specific text data.
	\item We propose an $n$-gram mixture language model, \texttt{graphmax}, which can be plugged into a global pre-trained LM for a scene-specific task. 
	\item We evaluate the performance of \texttt{graphmax} on text generation and machine translation tasks. The experimental results demonstrate that \texttt{graphmax} outperforms the traditional \texttt{softmax} in scene-specific text generation and macine translation.
\end{enumerate}

The rest of the paper is organized as follows. In Section 2, we review the literature relevant to our method and present the architecture of \texttt{graphmax} for text generation. In Section 3, we propose to optimize the proposed model with the projected gradient descent, and Section 4 elaborates on  the detailed algorithm and theoretical properties of \texttt{graphmax}.
The experimental results are presented in Section 5, and Section 6 concludes with some remarks.


\section{Methodology}
In this section, we will present the fundamental framework of \texttt{graphmax}. To establish a solid foundation, we commence by conducting a literature review on the topic of \texttt{softmax}.
\subsection{Literature Review on Softmax}
The \texttt{softmax} function is fundamentally important for many applications in machine learning. It produces a probability distribution across multiple categories, enabling models' prediction capability, e.g., recognizing a digital handwritten image or generating the next word in a text sequence. 
However, the traditional \texttt{softmax} has limitations when applied to high-dimensional scenarios, such as text generation \shortcite{yang2018breaking}. To address this issue, \citeA{yang2018breaking} proposed a mixture softmax function that improves the expressiveness of the \texttt{softmax} operator by replacing the single layer in the traditional \texttt{softmax} with an ensemble layer that has more parameters.
In language modeling, however, a word can have multiple meanings depending on the context, which means that a single word may lead to multi-sense embeddings. \shortciteA{miao2019kernelized} proposed a kernelized Bayesian softmax function that enhances the expressiveness of the \texttt{softmax} operator by incorporating a multi-sense kernel function. This approach provides a more flexible and expressive framework for modeling word senses and contexts in language modeling tasks.

\citeA{gao2017properties} provided a comprehensive summary and analysis of the properties of the \texttt{softmax} function using the convex analysis and monotone operator theory. They demonstrated that the \texttt{softmax} function is a monotone gradient map of the log-sum-exp function.
In text generation, the \texttt{softmax} function computes a probability distribution over the words in a dictionary and produces a high-dimensional sparse vector.  \citeA{martins2016softmax} proposed a sparse softmax function that returns a sparse posterior distribution, where the loss function of the sparse \texttt{softmax} is analogous to the logistic loss. 
However, the existing works fail to consider the relationships among all components of the sparse vector.
Overall, \citeA{gao2017properties} provided a valuable theoretical foundation for understanding the properties of the \texttt{softmax} function, while  \citeA{martins2016softmax} offered a practical solution for handling the sparsity issue associated with \texttt{softmax} in text generation tasks.

To model the concurrent relationships among words, we propose a GTV-regularized \texttt{softmax} function where GTV characterizes the corpus-specific knowledge and improves the local fluency in text generation. In contrast, the total variation (TV), proposed by \citeA{rudin1992nonlinear}, is a spatial regularity that penalizes signals with excessive and possibly spurious local detail, which has been widely applied in computer vision and signal processing \cite{lellmann2009convex,chambolle2010introduction,jia2019regularized,lellmann2011continuous}. For example, \citeA{lellmann2009convex} and \citeA{lellmann2011continuous} applied TV to multi-class image labeling, and \citeA{jia2019regularized} proposed a 2D TV-regularized U-net for image segmentation. 

The use of graphical techniques for signal processing has become increasingly popular in dealing with signals that have irregular structures, such as biological mechanisms, social networks, and citation networks. Recent studies by \citeA{chen2014signal} and \shortciteA{chen2015signal} have extended graphical signal recovery techniques to matrix completion \shortcite{liu2016manifold} and semi-supervised learning \shortcite{lv2022ssl}, where TV over a graph is imposed on the object of matrix completion. Through investigation of the relevance of TV and graph energy in graph signal classification,
\citeA{ahmed2017graph} found that TV is a compact and informative attribute for efficient graph discrimination while graph energy aims to quantify the complexity of the graph structure. These findings have important implications for understanding the relationship between signal processing and the underlying graph structure.
In a related study, \citeA{raguet2018cut} extended the cut-pursuit algorithm to the GTV regularization of functions with a separable nondifferentiable part. 

Many existing works on GTV are focused on undirected graphs, while there has been increasing interest in directed graphs \shortcite{shi2019skeleton}. In the context of text generation, the relationships between concurrent words are directed and weighted, which has motivated researchers to derive GTV over a weighted directed graph. 

\subsection{Conditional Text Generation } 
Deep LMs have been shown to be promising for conditional natural language generation with users' additional input. In controllable text generation tasks with different targets, different types of conditional constraints are often required, such as an image in image captioning \shortcite{anderson2017guided} or a style embedding vector in task-specific text generation \cite{ficlergoldberg2017controlling}. 

Traditional natural language generation methods rely on a large corpus, which is typically expensive to train. To address this issue, researchers have explored a new paradigm of pre-trained LMs, such as BERT \shortcite{devlin2019bert}, GPT-2 \cite{radford2019language}, and BART \shortcite{lewis2020bart,Koto2020FFCIAF,Chen2022TwophaseME,Vougiouklis2020PointAT}. By incorporating additional domain-specific codes \shortcite{keskar2019ctrl}, such as sentiment labels \shortcite{Dathathri2020Plug} or attribute vectors \cite{yu2021attributealignment}, the goal is to modify the pre-trained LM with little fine-tuning cost. For example, the plug-and-play language model (PPLM) \cite{Dathathri2020Plug} builds a user-specified bag-of-words classifier on top of GPT-2 to increase the likelihood of the target attribute. 
\shortciteA{koncelkedziorski2019text} proposed text generation from graph-based constraints, while it only inputs the knowledge graph embedding into a transformer-based model for text generation, which is an implicit way of using the graph. In contrast, our proposal explicitly regularizes the process of text generation by incorporating GTV over a weighted directed graph.

\subsection{The Softmax Function}
Given a context word sequence ${\bf w}_{1:t-1}=(w_1,\ldots,w_{t-1})$, at each decoder time step, the feature map of $(w_1,\ldots,w_{t-1})$ is denoted as $\mathbf{z} = \phi({\bf w}_{1:t-1})$, where $\phi(\cdot)$ is a well-trained large-scale LM, $\mathbf{z}=(z_1,\ldots,z_N)\in \mathbb{R}^N$, and $N$ is the cardinality of a dictionary. Typically, we use the \texttt{softmax} function to link $\mathbf{z}$ with the prediction of the next word $w_t$. From the perspective of optimization, the \texttt{softmax} function corresponds to the minimizer of the following problem\footnote{The object of the optimization is the conjugate function of the negative entropy $\langle \mathbf{x},\log \mathbf{x}\rangle$.},
\begin{equation}
	\begin{aligned}
		& \underset{\mathbf{x}}{\text{min}}
		& &-\langle \mathbf{x},\mathbf{z}\rangle + \langle \mathbf{x},\log \mathbf{x}\rangle \quad
		& \text{s.t.}
		& & \mathbf{1}^{\top}\mathbf{x}=1, 
	\end{aligned}
\label{eq:softmaxOpt}
\end{equation}
where $\mathbf{x}\in \mathbb{R}^N$ is a graph signal \cite{chen2014signal}, and $\mathbf{1}$ is an $N$-vector of all ones. The global minimizer of (\ref{eq:softmaxOpt}) is denoted by $\mathbf{x}^*$, whose $i$-th component is
\begin{equation}
	x^*_i = \frac{e^{z_i}}{\sum_{i=1}^N e^{z_i}},\quad i=1,\ldots,N,
\label{eq:softmax}
\end{equation}
where $z_i$ is the $i$-th component of $\mathbf{z}$.

\subsection{Graphmax: Graph Regularized Softmax}
\label{sec:graphsoftmax}
In text generation, an LM predicts the next word based on the history word sequence with a \texttt{softmax} function. Suppose the total number of words in a dictionary is $N$. An $N$-dimensional context representation vector $\mathbf{z}$ can be taken as the input for the \texttt{softmax} mapping as shown in Equation (\ref{eq:softmax}). By constructing a directed graph $G = (V, \mathbf{A}, \mathcal{C})$ of the words in the dictionary based on a large human corpus, we can take this graph $G$ as an elementary filtering operation that replaces a signal coefficient for each word with a weighted linear combination of coefficients at its neighboring nodes of $G$, where $V = \{v_i\}_{i=1}^N$ is the node set, $\mathbf{A}$ is an $N\times N$ weighted adjacency matrix, and $\mathcal{C}$ is the corpus that graph $G$ is built upon.

The smoothness of signals over $G$ can be quantified by the graph total variation (GTV),
\begin{equation*}
	\textrm{GTV} = \norm{ \mathbf{x} - \frac{1}{|\mu_{\textrm{max}}|}\mathbf{A}\mathbf{x} }_2^2,
\end{equation*}
where $\mu_{\textrm{max}}$ denotes the largest eigenvalue of the adjacency matrix $\mathbf{A}$. In practice, we can normalize the adjacency matrix $\mathbf{A}$ to make its maximum eigenvalue $\mu_{\textrm{max}}=1$ to simplify the computation. 
Specifically, for a node $v_i$, its outgoing edges have weights $1/{\rm deg}(v_i)$, where ${\rm deg}(v_i)$ is the out-degree of $v_i$ (the number of outgoing edges). Via matrix normalization, we set $\tilde{\mathbf{A}} = \mathbf{D}^{-1}\mathbf{A}$ \shortcite{schlichtkrull2018modeling,shi2019skeleton}, where $\mathbf{D}$ is a diagonal matrix with the $i$-th diagonal element
$\mathbf{D}_{ii}=\sum_j \mathbf{A}_{ij}+\epsilon$ and $\epsilon$ is a small number to avoid division by zero.
The normalization of the adjacency matrix ensures that the largest eigenvalue $\mu_{\textrm{max}}=1$.

By imposing GTV on the optimization problem in Equation (\ref{eq:softmaxOpt}), we obtain a graph-regularized \texttt{softmax},
\begin{equation}
	\begin{aligned}
		& \underset{\mathbf{x}}{\text{min}}
		& &-\langle \mathbf{x},\mathbf{z}\rangle + \langle \mathbf{x},\log \mathbf{x}\rangle + \lambda \|\mathbf{x} - \tilde{\mathbf{A}}\mathbf{x}\|_2^2 \\
		& \text{s.t.}
		& & \mathbf{1}^{\top}\mathbf{x}=1, \quad \mathbf{x},\mathbf{z}\in \mathbb{R}^N,
	\end{aligned}\label{eq:softmaxGTV}
\end{equation}
where $\tilde{\mathbf{A}}$ is a normalized adjacency matrix of the directed graph and $\lambda>0$ is a tuning parameter. We name the solution to (\ref{eq:softmaxGTV}) as \texttt{graphmax}.
The GTV regularizer in (\ref{eq:softmaxGTV}) forces ${\bf x}$ to be close to $\tilde{\mathbf{A}}\mathbf{x}$ where $\tilde{\mathbf{A}}$ is a (normalized) adjacency matrix of the directed graph that contains local knowledge from scene-specific corpus. Note that $\tilde{\mathbf{A}})$ is like a projection matrix which projects ${\bf x}$ to the space of $\tilde{\mathbf{A}})$.  Imagining $\tilde{\mathbf{A}}={\bf I}$, an identity matrix, the regularizer would not exist (equal to zero), which indicates the components of ${\bf x}$ are all independent so that no local information on scene-specific text is utilized. By forcing ${\bf x}$ to be close to $\tilde{\mathbf{A}}={\bf I}$, the local scene-specific knowledge would be incorporated in predicting the choice probabilities for the next word.


\section{Optimization}
\label{sec:4}
It is difficult to solve Equation (\ref{eq:softmaxGTV}) with traditional methods, such as the gradient descent or Lagrangian method. Instead, we propose to optimize it with the projected gradient descent detailed as follows.

\subsection{Projected Gradient Descent}
In the optimization problem of Equation (\ref{eq:softmaxGTV}), the feasible set $\Delta=\{\mathbf{x}| \mathbf{1}^{\top}\mathbf{x}=1, 0\leq x_i \leq 1\}$ is a probability simplex, which is a convex set. The graph regularized objective function is 
\begin{equation*}
	f(\mathbf{x}) = -\langle\mathbf{x},\mathbf{z}\rangle + \langle \mathbf{x},\log \mathbf{x} \rangle + \lambda \|\mathbf{x} - \tilde{\mathbf{A}}\mathbf{x}\|_2^2 \ ,
\end{equation*} 
which can be shown to be
strictly convex as follows. We examine the second-order condition of $f(\mathbf{x})$,
\begin{equation}\label{Hessian}
	\nabla^2 f(\mathbf{x}) = {\rm diag}\left(\frac{1}{\mathbf{x}}\right) + 2\lambda (\mathbf{I}-\tilde{\mathbf{A}})^{\top}(\mathbf{I}-\tilde{\mathbf{A}}).
\end{equation}
Because	$\mathbf{I}$ is an identity matrix and $\tilde{\mathbf{A}}$ is a normalized adjacency matrix of the directed graph $G$,  $(\mathbf{I}-\tilde{\mathbf{A}})$ is non-singular. That is, $(\mathbf{I}-\tilde{\mathbf{A}})\mathbf{x}=\mathbf{0}$ holds only if $\mathbf{x}=\mathbf{0}$. Hence, the matrix $(\mathbf{I}-\tilde{\mathbf{A}})^{\top}(\mathbf{I}-\tilde{\mathbf{A}})$ is positive definite, which leads to the conclusion that the Hessian matrix in (\ref{Hessian}) is positive definite 
as $\lambda >0$.

The optimization problem in Equation (\ref{eq:softmaxGTV}) can be solved by the projected gradient descent algorithm \shortcite{duchi2008efficient,MAL058}, which involves two steps: the gradient descent step and a subsequent projection of the gradient onto a probability simplex. Specifically, the two steps are given by 
\begin{equation}\label{eq:projectedGradientDescent}
	\begin{cases}
		\mathbf{a}^{t+1}\leftarrow \mathbf{x}^{t} - \alpha \nabla f(\mathbf{x}^{t}), & \text{gradient descent},\\
		\mathbf{x}^{t+1}\leftarrow \Pi_{\mathcal{C}}(\mathbf{a}^{t+1}), & \text{gradient projection},
	\end{cases}
\end{equation}
where $\alpha$ is a learning rate, $\mathbf{a}^{t+1}$ is an intermediate updated gradient at iteration $t+1$
and $\Pi_{\mathcal{C}}(\cdot)$ is the projection operator. 

\subsection{Projection onto Probability Simplex $\Delta$}
\label{sec:projectionDelta}
Given any closed set $\Delta \subset \mathbb{R}^N$, for any point $\mathbf{a}\in \mathbb{R}^N$ the projection operator $\Pi_{\Delta}(\cdot)$ is deﬁned as
\begin{equation*}
	\Pi_{\Delta}(\mathbf{a})=\argmin_{\mathbf{x}\in \Delta} \|\mathbf{x}-\mathbf{a}\|_2 .
\end{equation*}

In the case of $\Delta=\{\mathbf{x}| \mathbf{1}^{\top}\mathbf{x}=1, 0\leq x_i \leq 1\}$, the definition of the projection operation  $\Pi_{\Delta}(\cdot)$ can be reformulated as an optimization procedure,
\begin{equation}\label{eq:projectSimplex}
	\begin{aligned}
		\min_{\mathbf{x}} \quad & \|\mathbf{x}-\mathbf{a}\|_2\quad
		\textrm{s.t.} \quad & \sum_{i=1}^N x_i=1, \ \ \
		&x_i\geq0,   \\
	\end{aligned}
\end{equation}
which can be solved by the Lagrangian method \cite{duchi2008efficient}. By applying the standard KKT conditions for the Lagrangian of Equation (\ref{eq:projectSimplex}), the $i$-th component of the optimal $\mathbf{x}^*$ can be obtained as
\begin{equation}
	x_i^* = \max\{a_i+\gamma,0\}, \quad i=1,\ldots,N.
	\label{eq:graphsoftDefinition}
\end{equation}
where $\gamma$ can be computed from a vector of temporal variables $\mathbf{b}=(b_1,\ldots,b_N)\in \mathbb{R}^N$ and an integer $k$,
\begin{equation*}
	\gamma = \frac{1}{k}\left(1-\sum_{j=1}^k b_j\right).
\end{equation*}
The vector $\mathbf{b}=(b_1,\ldots,b_N)$ can be easily obtained by applying a sort function ${\rm sort}(\cdot)$ to the vector $\mathbf{a}\in \mathbb{R}^N$,
\begin{equation*}
	(b_1,\ldots,b_N) = {\rm sort}(\mathbf{a}),
\end{equation*}
such that $b_1 \leq \cdots \leq b_N$. 
The integer $k$, $1\leq k \leq N$, can be calculated from the sorted vector $\mathbf{b}$,
\begin{equation*}
	k=\argmax_{1\leq j \leq N}\left\{b_j+\frac{1}{j}\left(1-\sum_{i=1}^j b_i\right)>0\right\}.
\end{equation*}

\section{Computation and Theoretical Properties}
\label{sec:graphsoftmax_discussion}
In this section, we will outline the procedure of the \texttt{graphmax} algorithm and subsequently delve into a comprehensive analysis of its theoretical properties.
\subsection{Computational Complexicity}
Algorithm \ref{algorithm:1} outlines the procedure for the \texttt{graphmax} algorithm. In particular, step 3 computes the current gradient using gradient descent, and steps 4--7 correspond to the gradient projection $\Pi_{\Delta}(\cdot)$. The time complexity of the gradient projection onto the probability simplex is dominated by the cost of sorting the components of vector $\mathbf{a}$, which is $\mathcal{O}(N\log N)$. The maximum number of iterations is $T$, and thus the complexity of Algorithm \ref{algorithm:1} is $\mathcal{O}(TN\log N)$.

In practice, the value of $T$ is typically small. In our experiments, we set $T=20$ and both the learning rate and the convergence threshold as $10^{-4}$, which are reasonable choices for most applications.
\IncMargin{0.5em}
\begin{algorithm}[h] 
	\SetKwInOut{Input}{Input}
	\SetAlgoLined
	\Input{ Learning rate $\alpha$, maximum number of iteration steps $T$} 
	\BlankLine
	$\mathbf{x}^{1} \leftarrow \mathbf{0}$\;
	\While{$t<T$ 
 }{
		$\mathbf{a}^{t+1} \leftarrow \mathbf{x}^{t} - \alpha \nabla f(\mathbf{x}^{t})$ \;
		Sort $\mathbf{a}^{t+1}=(a^{t+1}_1,\ldots,a^{t+1}_N)$ as $(b^{t+1}_1,\ldots,b^{t+1}_N)$\;
		Find $k=\max\{1\leq j \leq N:b_j^{t+1}+(1-\sum_{i=1}^j b_j^{t+1})/j>0\}$\;
		Calculate $\gamma = (1-\sum_{i=1}^k b_j^{t+1})/k$\;
	Calculate $x_i^{t+1}=\max\{a^{t+1}_i+\gamma,0\}$\;
  If $\|\mathbf{x}^{t+1} - \mathbf{x}^{t}\|_2<10^{-4}$ then break;
	}
 \SetKwInOut{Output}{Output}
 \Output{ A near-optimal minimum ${\mathbf{x}}^*$ of the \texttt{graphmax}.}
	\caption{Projected Gradient Descent for \texttt{graphmax}.\label{algorithm:1}}
\end{algorithm}

\subsection{How \texttt{graphmax} Works}
\label{sec:howitworks}
Step 3 of Algorithm \ref{algorithm:1} serves as an interface between the pre-trained language model and the word concurrent graph. In this step, we calculate the gradient $\nabla f(\mathbf{x}^{t})$ of the objective function, as shown in Equation (\ref{eq:softmaxGTV}). The gradient consists of three distinct parts, 
\begin{equation}\label{eq:gradient}
    \nabla f(\mathbf{x}^{t}) = \underbrace{ -\mathbf{z}}_{\text{global knowledge from LM}} + \log \mathbf{x}^{t} + 1 + \underbrace{2\lambda (\mathbf{I}-\tilde{\mathbf{A}})^{\top}(\mathbf{I}-\tilde{\mathbf{A}}) \mathbf{x}^{t}}_{\text{local knowledge from scene-specific corpus}},
\end{equation}
where the first part $-\mathbf{z}$ is the decoder output of a pre-trained LM, such as GPT-2. As explained in Section 3.1, we have $\mathbf{z} = \phi({\bf w}_{1:t-1})$, where $\phi(\cdot)$ is the well-trained large-scale LM, so that $\mathbf{z}$ carries the global linguistic knowledge from the pre-trained LM. It is important to note that the traditional \texttt{softmax} only receives $\mathbf{z}$ as the input and produces a probabilistic distribution as indicated by Equation (\ref{eq:softmax}). The second part in (\ref{eq:gradient}), $\log \mathbf{x}^{t} + 1$, depends solely on $\mathbf{x}^{t}$.
The third part in (\ref{eq:gradient}), $2\lambda (\mathbf{I}-\tilde{\mathbf{A}})^{\top}(\mathbf{I}-\tilde{\mathbf{A}}) \mathbf{x}^{t}$, is derived from the word concurrent graph, which is constructed from a scene-specific corpus (as depicted in Figure \ref{fig:demoGraph}). This part represents the local knowledge relevant to the target task. The local knowledge is modeled as a second-order polynomial of the normalized adjacency matrix, denoted by $\tilde{\mathbf{A}}$. It is worth noting that $\tilde{\mathbf{A}}$ and $\tilde{\mathbf{A}}^2$ correspond to 2-gram and 3-gram models, respectively. Therefore, the \texttt{graphmax} with $(\mathbf{I}-\tilde{\mathbf{A}})^{\top}(\mathbf{I}-\tilde{\mathbf{A}})$ can be viewed as a {(2,3)}-gram mixture model with a penalty of $2\lambda$.

Equation (\ref{eq:gradient}) demonstrates that the local knowledge and the global pre-trained LM operate in a plug-in mode. This property allows the local knowledge to be easily integrated into any pre-trained LM using the proposed \texttt{graphmax}. Unlike traditional conditional natural language generation models that modify $\mathbf{z}$ with additional inputs, \texttt{graphmax} does not alter $\mathbf{z}$. Therefore, there is no need to fine-tune the pre-trained model. To perform task-specific text generation, we can simply attach the \texttt{graphmax} module with local knowledge to a global pre-trained model.

\subsection{Theoretical Properties}
For the \texttt{softmax} defined in Equation (\ref{eq:softmax}), we include the following proposition for completeness and comparison.

\begin{proposition}
\cite{martins2016softmax} Let $\mathbf{z}=(z_1,\ldots,z_N)\in \mathbb{R}^N$ be a feature map vector. If $z_i\leq z_j$, then the following inequalities hold,
\begin{align*}
     \texttt{softmax}_j(\mathbf{z}) - \texttt{softmax}_i(\mathbf{z}) & \geq 0, \\
    \texttt{softmax}_j(\mathbf{z}) - \texttt{softmax}_i(\mathbf{z}) & \leq \frac{1}{2}( z_j -  z_i),
\end{align*}
where $\texttt{softmax}_i(\mathbf{z})$ denotes the $i$-th component of $\texttt{softmax}(\mathbf{z})$.
\end{proposition}
The proof can be referred to \citeA{martins2016softmax}.
The proposed \texttt{graphmax} possesses a similar property.
\begin{proposition}
For a  feature map vector $\mathbf{z}=(z_1,\ldots,z_N)\in \mathbb{R}^N$, if $z_i\leq z_j$, then the following inequalities hold,
\begin{align*}
     \texttt{graphmax}_j(\mathbf{z}) - \texttt{graphmax}_i(\mathbf{z}) & \geq 0, \\
    \texttt{graphmax}_j(\mathbf{z}) - \texttt{graphmax}_i(\mathbf{z}) & \leq z_j -  z_i,
\end{align*}
where $\texttt{graphmax}_i(\mathbf{z})$ denotes the $i$-th component of $\texttt{graphmax}(\mathbf{z})$. 
\end{proposition} 
The proof of Proposition 2 is provided in Appendix \ref{sec:appendix_B}. This proposition shows that the \texttt{graphmax} function shares similar properties with the \texttt{softmax} function. As is well-known, the \texttt{softmax} function maps a vector of real numbers to a probability distribution over those numbers in a consistent manner, such that larger numbers correspond to larger probabilities. By comparing the two propositions, we observe that not only does the \texttt{graphmax} function maintain this consistency but it also has an upper bound.

\section{Experiments}

\label{sec:experiements}
\paragraph{Experimental Settings} All models are implemented using Pytorch 1.7 on an Intel(R) Xeon(R) CPU E5-2680 v4 2.40GHz, Tesla K80 GPU, and 128G memory, based on the Ubuntu 16.04 platform. The parameters in Algorithm \ref{algorithm:1} are specified as follows: The maximum number of iteration steps for the \texttt{graphmax} algorithm is set as $T=20$, the learning rate $\alpha$ is $10^{-4}$, and the convergence threshold is $10^{-4}$. Specifically, the iteration process stops when the $l_2$ norm of the difference between consecutive iterations, $\|\mathbf{x}^{t+1}-\mathbf{x}^{t}\|_2$, is less than $10^{-4}$. The reported results are averaged over 100 independent runs.

\paragraph{Datasets and Baseline Methods}
We choose the Amazon \cite{zhang2015character} and Yelp \cite{mcauley2013hidden,zhang2015character} datasets for general text generation, WMT'16 \shortcite{bojaretal2016findings} and WMT'21 corpora \shortcite{tran2021facebook} for machine translation. 
The baseline methods are chosen as follows,
\begin{itemize}
    \item For text generation: GPT-2 \cite{radford2019language}, CTRL (Conditional Transformer Language model) \cite{keskar2019ctrl}, PPLM (Plug-and-Play LM) \cite{Dathathri2020Plug}, and OPT (Open Pre-trained Transformer) \shortcite{zhang2022opt}\footnote{https://huggingface.co/facebook/opt-125m}; LLAMA2 \shortcite{touvron2023llama}.
    \item For machine translation: BART \cite{lewis2020bart} and mBART \shortcite{liu2020multilingualdenoising}.
\end{itemize}
Generally, we set the baselines at their best configurations as reported in the respective papers. For examples, we set PPLM with the sentiment label, KL-scale 0.1, GM-scale 0.95, and step size 0.05. We assign CTRL with a review domain control code in the corresponding experiments.

\begin{figure}[h]
	\centering
	\includegraphics[width=8cm]{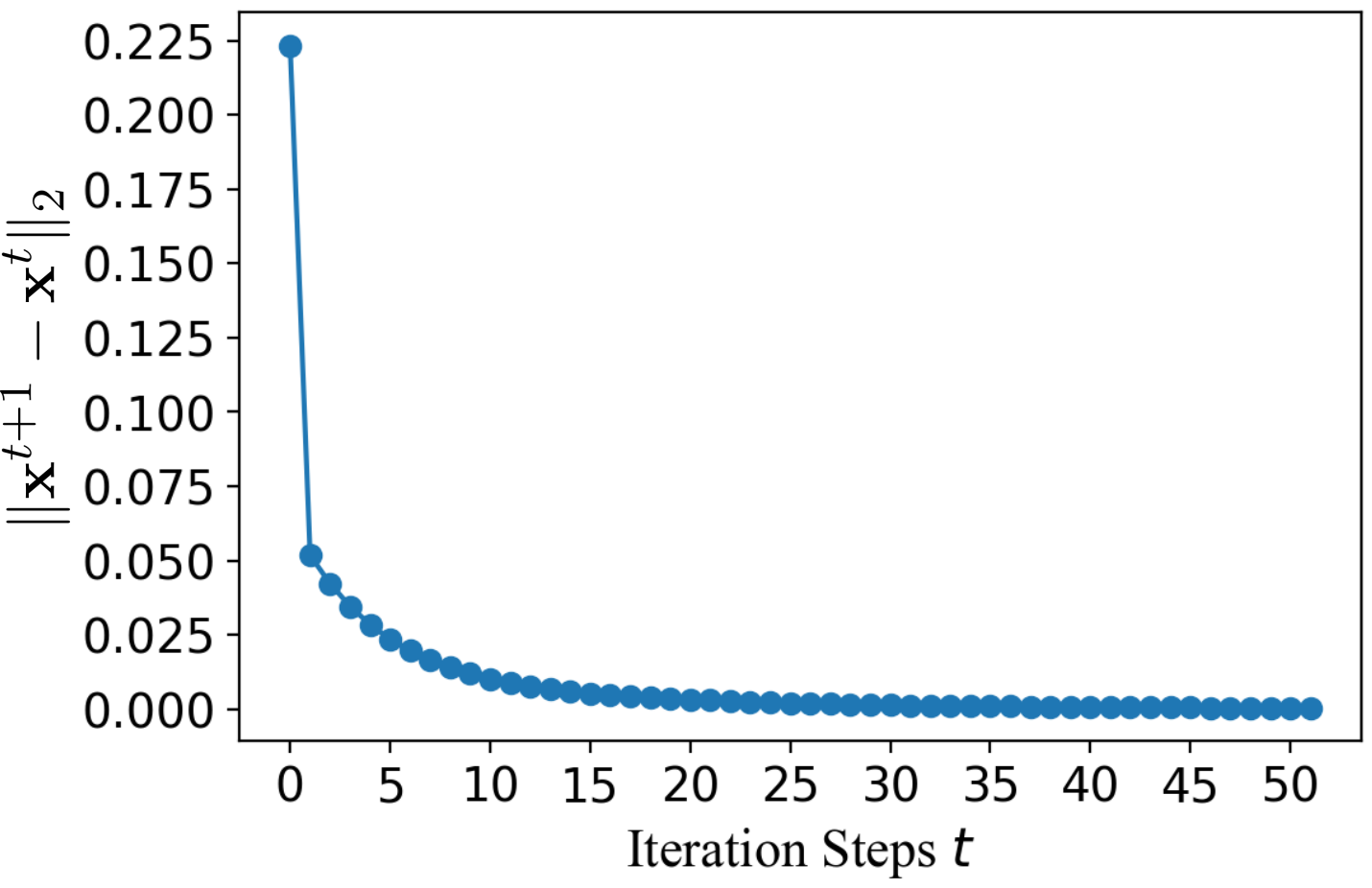}
	\caption{The visualization process of Algorithm \ref{algorithm:1}, i.e., $\|\mathbf{x}^{t+1} - \mathbf{x}^{t}\|_2$ versus iteration steps $t$.}
	\label{fig:delta_T}
\end{figure}

\begin{table}[!ht]
\scalebox{0.95}{
\begin{tabular}{l|p{14cm}}
\hline
GPT-2 &  {\em \textbf{You do not want to read this book if you don't want to change.} {Shepherd Tebbit Reading is business. It's business in the eyes of God. Hertha Wise Sorry to end this, but this has been a hunch for a couple of months now. See more posts on wikipedia.com. Also read Martin Walker's The Last Dance.} Updated 5:18 PM} \\\hline
CTRL & {\em \textbf{You do not want to read this book if you don't want to change.} It is not a guide to your own life, it is not an answer to your own questions. It is a warning to others to take the time to read this book. It is not an easy book to follow, and you will be left with a lot of questions that are going to be difficult to answer. This book is not about being afraid of change. It is about being a true believer. I think you are probably going to find a lot of mistakes, and that's okay. In this book, we are going to learn a lot and learn how to change our life. That means that you are going to find a lot of happiness. }\\ \hline
PPLM & {\em \textbf{You do not want to read this book if you don't want to change.} {The first part is} {an explanation of} what it is like to play with a computer and then to get your head around it. {You might just want to play with a computer, or you might want to use a calculator or a phone to get a picture of what to do. I really think it's important that you understand the basics of your game.} }\\ \hline
OPT & {\em \textbf{You do not want to read this book if you don't want to change.} This is a very inspiring book, which encourages you to change your life and become a much better person. If you're unhappy with the way you are, you should read this book! 
The author describes in the book several ways to improve yourself and your personality.}\\ \hline
New  &  {\em \textbf{You do not want to read this book if you don't want to change.} {Blossom flows, thickly}. Whatever you say it has become pretty clear. { You are learning from the book and finding things} that you must learn. {You are learning from the criticism, the griping, and the washing up. What you learn and remember about it is that it is what we learned from it}, {and so what we learn will help us get back into this new energy of determination that helped us overcome the most difficult elements of battle. Years ago during our fighting back before we even got a chance to fight back, we will never see a great American.}}\\ \hline
\end{tabular}
}
\caption{Comparisons of book reviews generated by GPT-2, CTRL, PPLM, OPT and GPT-2+\texttt{graphmax} (denoted as `New') on generating product reviews, with the local graph of the \texttt{graphmax}  constructed from the Amazon review corpus.}
\label{tab:compares_book_review}
\end{table}

\begin{table}[tb]
\renewcommand\arraystretch{1.50}
\centering
\scalebox{1.0}{
\begin{tabular}{lcccc}
\toprule
Methods & BLEU-2 &  BLEU-3 & BLEU-4 & BLEU-5\\ \hline
 GPT-2  & $0.983_{\pm  0.017} $ & $0.919_{\pm  0.024}$ &$ 0.788_{\pm  0.041} $ &$ 0.615_{\pm  0.021} $ \\
 GPT-2+\texttt{graphmax} & $\textbf{0.989}_{\pm  0.021}$  & $\textbf{0.945}_{\pm  0.062}  $ & $\textbf{0.832}_{\pm  0.026} $ &$ \textbf{0.660}_{\pm  0.015} $\\
\hline
CTRL &$ 0.965_{\pm  0.047}$  & $0.912_{\pm  0.031} $ &$ 0.832_{\pm  0.017}$  &$ 0.584_{\pm  0.014} $\\
CTRL+\texttt{graphmax} & $\textbf{0.971}_{\pm  0.064}$  & $\textbf{0.947}_{\pm  0.049} $ &$ \textbf{0.854}_{\pm  0.022}$ &$ \textbf{0.589}_{\pm  0.031} $\\
\hline
PPLM & $ 0.988_{\pm  0.028}$  &$ 0.935_{\pm  0.032} $ &$ 0.761_{\pm  0.049} $ &$ 0.631_{\pm  0.051} $\\
PPLM+\texttt{graphmax} & $\textbf{0.991}_{\pm  0.034} $ & $\textbf{0.946}_{\pm  0.084} $ & $\textbf{0.812}_{\pm  0.049 } $ &$ \textbf{0.673}_{\pm  0.021} $\\
\hline
OPT  & $0.995_{\pm  0.045} $ & $0.967_{\pm  0.049} $ & $0.876_{\pm  0.038} $ &$ 0.788_{\pm  0.011} $\\
OPT+\texttt{graphmax} & $\textbf{0.998}_{\pm  0.019} $ & $\textbf{0.978}_{\pm  0.022}$  &$ \textbf{0.912}_{\pm  0.011 }$ &$ \textbf{0.817}_{\pm  0.032} $\\ 
\hline
LLAMA2  & $\textbf{0.988}_{\pm  0.034} $ & $0.987_{\pm  0.067} $ & $0.910_{\pm  0.038} $ &$ 0.845_{\pm  0.011} $\\
LLAMA2+\texttt{graphmax} & $\textbf{0.998}_{\pm  0.026} $ & $\textbf{0.988}_{\pm  0.045}$  &$ \textbf{0.922}_{\pm  0.011 }$ &$ \textbf{0.867}_{\pm  0.032} $\\
\bottomrule
\end{tabular}
}
\caption{The BLEU scores of GPT-2, CTRL, PPLM, OPT, LLAMA2, and the proposed methods on generating product reviews ($\pm$ standard deviation), averaged over five cross-validation folds. 
The local graph is constructed with the Amazon corpus, and the best results are highlighted in boldface. }
\label{tab:bleu_amazon}
\end{table}

\paragraph{Evaluation Metrics}
We employ a hybrid approach for performance evaluation, combining both human and automatic metrics. Specifically, we utilize two well-known automatic metrics: BLEU (Bilingual Evaluation Understudy) \shortcite{papineni2002bleu} and perplexity (PPL), for assessing the quality of text generation and machine translation.

BLEU counts the matches of $n$-grams in the generated text with $n$-grams in the reference text. For instance, a 1-gram comparison would consider each individual token, whereas a 2-gram comparison would assess every pair of words. We employ BLEU-2, BLEU-3, BLEU-4, BLEU-5 scores to measure the $\{2,3,4,5\}$-gram matching between the generated text and reference text respectively.

Another important metric is the sentence fluency, which depicts the naturalness of human expressions. However, assessing the fluency of a natural language generation system is challenging \cite{kann2018sentence}.
Therefore, we use human testers to evaluate the fluency of the generated text as a supplement.

\paragraph{Graph Construction}
We conceptualize a corpus as a directed graph, wherein the vertices represent individual words, and the edges denote the sequential relationship between two adjacent words in a sentence. The direction of each edge is determined by the word link of a 2-gram. We then tally the frequency of the 2-grams appeared in our corpus as the weight of the corresponding edge. This graphical representation intuitively captures the natural flow of expression.

To construct these graphs, we utilize datasets employed in our experimental analysis. For instance, we construct the word concurrent graph by leveraging datasets from food/restaurant reviews or e-commerce reviews sourced from Yelp. These datasets exhibit distinct characteristics that diverge from the typical structure of general text. Notably, the online review language tends to incorporate more colloquial phrases, shorter sentences, and internet slangs. These unique features of review text present a convenient opportunity to juxtapose the local fluency of our proposed model with that of other approaches.

The concurrent graph is constructed based on the statistics of 2-grams, which corresponds to a 2-gram model. However, this construction can be easily extended to an $n$-gram model by utilizing a higher-order adjacency matrix, denoted as $\tilde{\mathbf{A}}^{n-1}$.
As discussed in Section \ref{sec:graphsoftmax_discussion}, the proposed \texttt{graphmax} acts as a $\{2,3\}$-gram mixture model when the input is a first-order normalized adjacency matrix $\tilde{\mathbf{A}}$. For an $n$-order input matrix $\tilde{\mathbf{A}}^n$, \texttt{graphmax}  models an $\{n+1, n+2, \ldots, 2n+1\}$-grams mixture.
In practice, we can even augment the input matrix by replacing $\tilde{\mathbf{A}}$ with $\tilde{\mathbf{A}}+\tilde{\mathbf{A}}^2+ \cdots +\tilde{\mathbf{A}}^n$, which yields a more complex $n$-grams mixture model. This approach further captures higher-order word dependencies and enhances the ability to model complex language structures.

\paragraph{Convergence of \texttt{graphmax}}
The core part of our model is the projection gradient descent, which is used to calculate the GTV regularized \texttt{softmax}. Unlike the traditional \texttt{softmax}, which has a closed form solution, the proposed \texttt{graphmax} needs to be computed iteratively.  Fortunately, the computational cost of this iterative procedure is still affordable. 
Taking the largest dataset Amazon as an example, the corresponding directed graph contains a total of $N=50527$ words, and the adjacency matrix $\tilde{\mathbf{A}}$ is of size $50527 \times 50527$.
Figure \ref{fig:delta_T} shows that
the \texttt{graphmax} converges in 20 iteration steps, where the X-axis represents the iterative steps $t$ and the Y-axis is the $l_2$ gap between two adjacent iterations $\|\mathbf{x}^{t+1} - \mathbf{x}^{t}\|_2$. Table \ref{tab:timeStat} presents a comparison of the average time required for inferring the next word in five experimental scenarios, between LLAMA2 and LLAMA2+\texttt{graphmax}. It is worth noting that the time cost associated with \texttt{graphmax} is approximately 15--17 times higher than that of the original LLAMA2 implementation, which is consistent with the findings derived from the analysis of time complexity.

\begin{table}[]
\centering
\begin{tabular}{@{}lccccc@{}}
\toprule
 LLAMA2& Prompt 1 & Prompt 2 & Prompt 3 & Prompt 4 & Prompt 5 \\ \midrule
without \texttt{graphmax}    & 0.092    & 0.085    & 0.097    & 0.073    & 0.103    \\
with \texttt{graphmax} & 1.21     & 1.41     & 1.43     & 1.14     & 1.24     \\ \bottomrule
\end{tabular}
\caption{Comparison of the average inference time (in seconds), with the base model LLAMA2 \shortcite{touvron2023llama} and the local corpus  Amazon.}
\label{tab:timeStat}
\end{table}

\subsection{General Text Generation}
\paragraph{E-Commerce Product Review Generation}
In the e-commerce product review generation experiment, we construct a graph with 4 million product reviews in the Amazon corpus  \cite{zhang2015character}. 
Table \ref{tab:compares_book_review} shows representative book reviews generated by the four baseline methods (GPT-2, CTRL, PPLM, OPT) as well as the proposed \texttt{graphmax} incorporated into GPT-2 (GPT-2+\texttt{graphmax}). 
From the perspective of narrative logic, PPLM, GPT-2, OPT, and our method perform better than CTRL. The sentences generated by CTRL often appear as paraphrases of a start sentence. On the other hand, GPT-2, OPT, and our method elaborate more on the topic and engage in further discussions.
The book review generated by GPT-2 contains extensive information, such as another book's title, people related to this book, etc., while such abundance of multi-source information can be distracting. Similarly, although OPT may highly recommend a book, the basis for such a recommendation is not attributed to the book's content.
In contrast, the proposed GPT-2 with \texttt{graphmax} attempts to gradually persuade the reader by focusing on the book's content. It utilizes sentences of varying lengths,
which resembles genuine reviews.

Table \ref{tab:bleu_amazon} shows the BLEU scores of the text generated by the baseline methods and our model. The relative improvement of GPT-2 with  \texttt{graphmax} over the baselines is more significant in the BLEU-5 metric than the BLEU-2 metric. 

\paragraph{Food Review Generation} This experiment constructs a graph based on the Yelp food review \cite{mcauley2013hidden,zhang2015character} dataset, which contains more than 700,000 food reviews. 

\begin{table}[!ht]
\scalebox{0.95}{
\begin{tabular}{l|p{14cm}}
\hline
GPT-2 &  {\em \textbf{Now I know why people review this place so highly}, it's even ridiculous that a local show will automatically get chosen as the greatest movie in the annals of writing. {I mean even though it's written by the ones I love in all these years! Just take a glance.} How many times have you seen that absolutely 100 dollar DVD and read the comments? That only proves that you are as mature as I am and as invested in The Cove as I am in Denzel Washington. Ridiculous. But I couldn't get a dentist just looking at them and thinking "well, the worst thing that could happen in a movie is the less you have to do it.
} \\ \hline
CTRL & {\em \textbf{Now I know why people review this place so highly}. {This is a place for good food, good music and amazing people.} I was here for an informal "crawl" on a Thursday night, and I saw a young woman and some young guys in their late 20s, {all sitting on the bench, drinking beer, with their feet dangling down in a pile of trash.} {We had been to a lot of great places here, but nothing beats a little party on an old-fashioned Saturday night and getting in to the groove and making some good friends.}}\\ \hline
PPLM & {\em \textbf{Now I know why people review this place so highly}.  {There's always a lot of reviews here for anything, but it's not a place where we're talking about an interesting place.} But I'm not sure.  This article was originally published at the University of Massachusetts Press. }\\ \hline
OPT&  {\em \textbf{Now I know why people review this place so highly}, Now I know why people review this place so highly, it's because it's a great place to stay. I've been staying here for years and I've never been disappointed. The rooms are clean and comfortable, the staff is friendly and helpful, and the location is great. I've stayed here for business and pleasure. }\\ \hline
New  &  {\em \textbf{Now I know why people review this place so highly}, for any reason—shear interest, get bored, try new things. However, I've been coming here for a few weeks now, {and when I first opened the door and saw this bar, my Roast was completely filled with treats, salads, and soul food}. I really enjoyed everything I've had in {here} for the past several days. {The food came out} in such a timely fashion that I stopped and have been {keeping an eye on the Trolley Room Lounge bar} (with its lights on) a little since I finished my morning. {The food is very good, as I've been ordering fresh baked foods since my college days, I still just couldn't stop with homemade sweet potatoes. I thought the bakery was pretty good}. {Overall, the atmosphere was clean, laid back, and friendly, and it was friendly.} }\\ \hline
\end{tabular}
}
\caption{Comparisons of restaurant reviews generated by GPT-2, CTRL, PPLM, OPT and GPT-2+\texttt{graphmax} (denoted as `New'), with the local graph constructed by the Yelp corpus.}
\label{tab:compares_examples_food}
\end{table}

\begin{table}[t]
\renewcommand\arraystretch{1.50}
\centering
\scalebox{1.0}{
\begin{tabular}{lcccc}
\toprule
Methods& BLEU-2 &  BLEU-3 & BLEU-4 & BLEU-5\\ \hline
 GPT-2  & $0.992_{\pm  0.044} $ & $0.931_{\pm  0.034}$ &$ 0.782_{\pm  0.032} $ &$ 0.590_{\pm  0.065} $ \\
 GPT-2+\texttt{graphmax} & $\textbf{0.994}_{\pm  0.038}$  & $\textbf{0.953}_{\pm  0.022}  $ & $\textbf{0.834}_{\pm  0.024} $ &$ 0.658_{\pm  0.008} $\\
\hline
CTRL &$ \textbf{0.989}_{\pm  0.047}$  & $0.912_{\pm  0.031} $ &$ 0.743_{\pm  0.017}$  &$ \textbf{0.607}_{\pm  0.014} $\\
CTRL+\texttt{graphmax} & $0.971_{\pm  0.054}$  & $\textbf{0.947}_{\pm  0.039} $ &$ \textbf{0.854}_{\pm  0.023}$ &$ 0.589_{\pm  0.041} $\\
\hline
PPLM & $ \textbf{0.992}_{\pm  0.32}$  &$ 0.941_{\pm  0.013} $ &$ 0.792_{\pm  0.049} $ &$ 0.621_{\pm  0.026} $\\
PPLM+\texttt{graphmax} & $0.991_{\pm  0.014} $ & $\textbf{0.946}_{\pm  0.033} $ & $\textbf{0.812}_{\pm  0.019 } $ &$ \textbf{0.673}_{\pm  0.034} $\\
\hline
OPT  & $0.997_{\pm  0.013} $ & $0.978_{\pm  0.031} $ & $0.845_{\pm  0.019} $ &$ 0.713_{\pm  0.021} $\\
OPT+\texttt{graphmax} & $\textbf{0.998}_{\pm  0.029} $ & $\textbf{0.985}_{\pm  0.042}$  &$ \textbf{0.876}_{\pm  0.009 }$ &$ \textbf{0.723}_{\pm  0.012} $\\ 
\hline
LLAMA2  & $\textbf{0.998}_{\pm  0.007} $ & $0.995_{\pm  0.078} $ & $\textbf{0.889}_{\pm  0.039} $ &$ 0.843_{\pm  0.062} $\\
LLAMA2+\texttt{graphmax} & $\textbf{0.998}_{\pm  0.029} $ & $\textbf{0.996}_{\pm  0.012}$  &$ 0.887_{\pm  0.034 }$ &$ \textbf{0.876}_{\pm  0.016} $\\ 
\bottomrule
\end{tabular}
}
\caption{The BLEU scores of GPT-2, CTRL, PPLM, LLAMA2, and OPT with or without  \texttt{graphmax} on generating restaurant reviews ($\pm$ standard deviation), averaged over five cross-validation folds. 
}
\label{tab:bleu_yelp}
\end{table}

Table \ref{tab:compares_examples_food} presents examples of the restaurant reviews generated by the baselines and the proposed model. 
From the view of the narrative logic, we observe that GPT-2 explores too much information, making it difficult to capture the main story-line ``a place''. In contrast, CTRL, PPLM, OPT, and our GPT-2+\texttt{graphmax} method can always make the text focused on the core story. 
We also see that OPT appears to resemble a blog post authored by a seasoned food enthusiast, utilizing a formal tone of expression rather than an internet review.
In contrast, the proposed method (GPT-2+\texttt{graphmax}) prefers a freestyle of text that is much closer to a real e-commerce review written by human users. In contrast, our model tends to use shorter phrases, such as the ``{\em shear interest}'', ``{\em get bored}'', ``{\em try new things}'', ``{\em laid back}'', and even a repetitive style of a statement, such as ``{\em and friendly, and it was friendly}'', which are more similar to real food/restaurant reviews on e-commerce sites. In other words, the fluency of the generated text approaches that of real reviews. The comparison between GPT-2 and our method (GPT-2+\texttt{graphmax}) suggests that the local knowledge obtained in Equation (\ref{eq:gradient}) is indeed important in text generation.

We report the BLEU-2 to BLEU-5 scores \cite{papineni2002bleu} in Table \ref{tab:bleu_yelp} to evaluate the overall performance for the food review generation. We set $\lambda=1.0$. 
The BLEU-2 to BLEU-5 scores on generating long food reviews using the models with \texttt{graphmax} (+\texttt{graphmax}) are consistently better than PPLM, CTRL, OPT, and GPT-2 without \texttt{graphmax}. For baseline LLAMA2, \texttt{graphmax} achieves better BLEU-3 and BLEU-5 but worse BLEU-4 than LLAMA2 without \texttt{graphmax}. 

\begin{figure}[!ht]
	\centering
	\includegraphics[width=14cm]{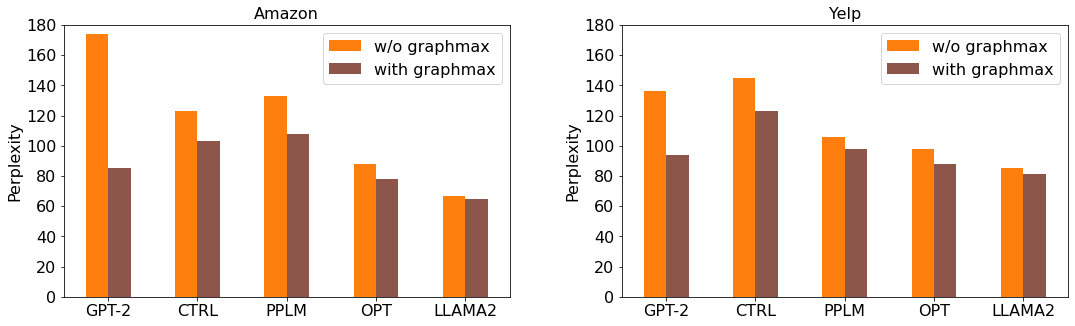}
\caption{Comparisons of perplexity using GPT-2, CTRL, PPLM, OPT, and LLAMA2 with or without (w/o) \texttt{graphmax} on generating product reviews,
with the local graph  constructed from the Amazon and Yelp corpuses.}
	\label{fig:visualizationPPL}
\end{figure}

The left panel of Figure \ref{fig:visualizationPPL} shows the perplexity (PPL) of the generated text by the competing methods and our model. The local graph is constructed with the Amazon corpus. Different from BLEU, perplexity is relatively low when the contribution of the regularized local knowledge (with \texttt{graphmax}) is small.
The right panel of Figure \ref{fig:visualizationPPL} summarizes the perplexity of generated food reviews. The results indicate that local knowledge benefits the improvement of perplexity for all the baseline models.

\begin{figure}[!ht]
	\centering
	\includegraphics[width=14cm]{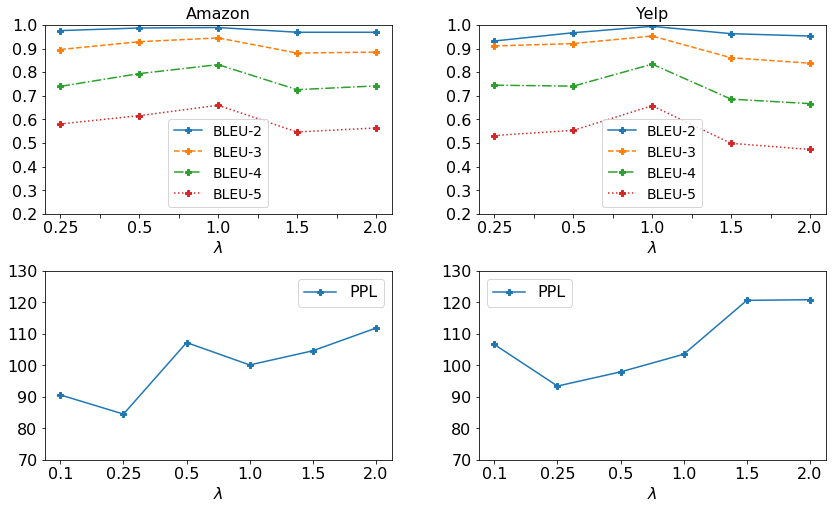}
	\caption{Searching $\lambda$ using GPT-2 with the \texttt{graphmax} approach, considering both BLEU and perplexity as evaluation metrics. The results are presented in two panels: the upper panel shows the evaluation from a BLEU perspective, while the bottom panel presents the evaluation based on perplexity. The local graph used for parameter searching is constructed from the Amazon corpus in the left panels and the Yelp corpus in the right panels.}
	\label{fig:lambda}
\end{figure}

\paragraph{Choice of $\lambda$}
The penalty parameter $\lambda$ for the GTV in Equation (\ref{eq:softmaxGTV}) plays an important role in our model. A larger value of $\lambda$ indicates that \texttt{graphmax} (local scene-specific knowledge) contributes more than the LM (global knowledge) on predicting the next word.
We choose $\lambda$ by taking an overall consideration of the BLEU and perplexity, because these two metrics evaluate our model from different perspectives. The experimental results indicate that the outcomes of BLEU and perplexity concerning $\lambda$ are not consistently aligned. Specifically, as depicted in the top 
panels of Figure \ref{fig:lambda}, BLEU attains the optimal value when $\lambda=1.0$, whereas perplexity generally favors a smaller $\lambda$, as demonstrated in the bottom 
panels of Figure \ref{fig:lambda}. Given our goal of training a model for generating scene-specific text, BLEU is more closely aligned with our objective than perplexity. Consequently, we 
select $\lambda=1.0$ as the default value.

\begin{figure}[!ht]
	\centering
	\includegraphics[width=12cm]{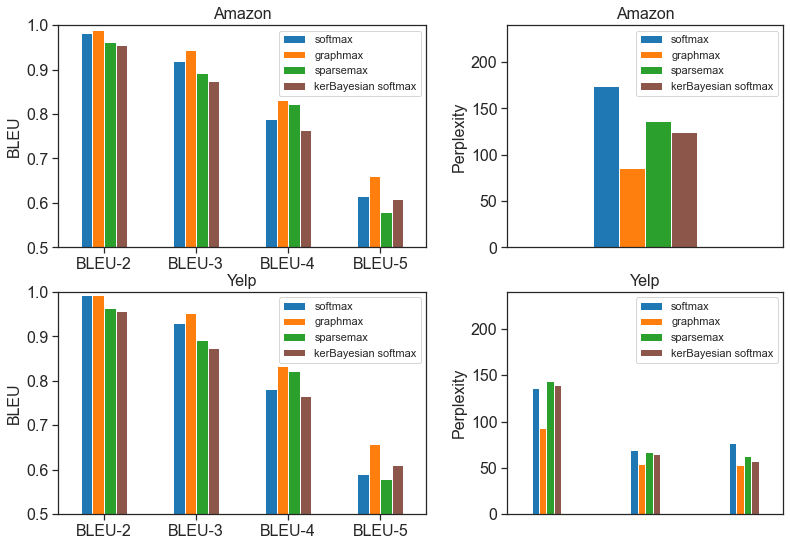}
	\caption{Comparisons of \texttt{graphmax} with \texttt{softmax}, \texttt{sparsemax}, and kernel Bayesian (kerBayesian) \texttt{softmax}  on review generation.  }
	\label{fig:4softmax}
\end{figure}

\paragraph{Comparisons with Other \texttt{softmax} Functions }
In this experiment, we compare the proposed \texttt{graphmax} with the original \texttt{softmax}, \texttt{sparsemax} \cite{martins2016softmax}, and kernel Bayesian \texttt{softmax}  (kerBayesian \texttt{softmax}) \cite{miao2019kernelized} functions on food and e-commerce review generation. As shown in Figure \ref{fig:4softmax}, 
the \texttt{graphmax} achieves the best performances on both the BLEU and perplexity metrics.

\paragraph{Visualization of $\mathbf{x}$}
\begin{figure}[!ht]
	\centering
	\includegraphics[width=11cm]{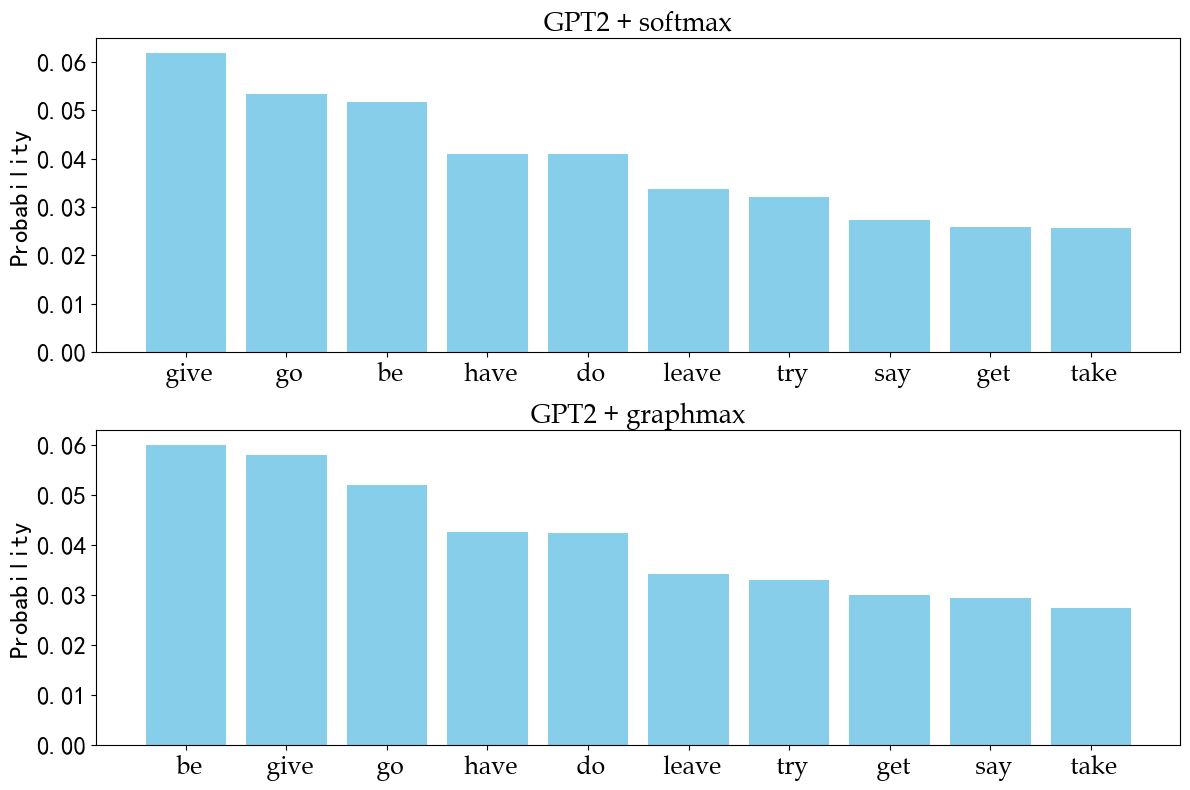}
	\caption{Visualization of the top 10 words respectively recommended by GPT2+\texttt{softmax} and GPT2+\texttt{graphmax} in the context ``{\em I was so disappointed by this, and at this stage I was going to $\cdots$ }''.}
	\label{fig:visualizationX}
\end{figure}

We visualize the vector calculated by the proposed \texttt{graphmax} and compare it with the corresponding results of the traditional \texttt{softmax}. 
In the context ``{\em I was so disappointed by this, and at this stage I was going to }$\cdots$'', we visualize the top 10 next words respectively recommended by GPT2+\texttt{softmax} and GPT2+\texttt{graphmax}. As shown in Figure \ref{fig:visualizationX}, the recommended lists of words by the two methods are slightly different.
The next word of the aforementioned context predicted by \texttt{softmax} is ``{\em give}'', while that predicted by \texttt{graphmax} is ``{\em be}''. 

\subsection{Machine Translation}
To evaluate the performance of \texttt{graphmax} in machine translation, we conduct both monolingual and multilingual machine translations.
We set the parameter $\lambda= 1.0$ in the following experiments.
\paragraph{Monolingual Machine Translation}
We use BART \cite{lewis2020bart} as a base model and link it to the proposed \texttt{graphmax}. We construct a local style graph with the target language of the WMT'16 corpus, and then plug the local word preference graph into BART to improve machine translation. Table \ref{tab:machinetrans} shows translation results on both Romanian $\rightarrow$ English and English $\rightarrow$ Romanian directions.

\begin{table}[]
    \centering
    \begin{tabular}{lcc}
        \toprule
      Methods   & Romanian$\rightarrow$English & English$\rightarrow$Romanian \\
        \midrule
        BART & 0.363 &  0.325 \\
        BART+\texttt{graphmax} & \textbf{0.372} & \textbf{0.335} \\
        \bottomrule
    \end{tabular}
    \caption{The BLEU scores of BART and BART+\texttt{graphmax} on monolingual machine translation.}
    \label{tab:machinetrans}
\end{table}

\paragraph{Multilingual Machine Translation}
As a baseline, we utilize mBART \cite{liu2020multilingualdenoising}, a multilingual denoising pre-trained model for multilingual translation tasks. We incorporate \texttt{graphmax} into mBART and evaluate the performance on the WMT'21 corpus \cite{tran2021facebook}, which comprises translations from English to Czech (Cz), German (Ge), Hausa (Ha), Icelandic (Ic), Japanese (Ja), Russian (Ru), and Chinese (Ch), as shown in Table \ref{tab:machinetransMultipleLan}. The upper panel of the table presents the BLEU scores for translating English into seven languages, while the lower panel displays the inverse translations. Our results demonstrate that mBART+\texttt{graphmax} surpasses the performance of the baseline model in most translation tasks.

\begin{table}[]
\setlength{\tabcolsep}{1.4mm}
\begin{tabular}
{@{}lccccccc@{}}
\toprule
Methods & En$\rightarrow$Cz & En$\rightarrow$Ge & En$\rightarrow$Ha & En$\rightarrow$Ic & En$\rightarrow$Ja & En$\rightarrow$Ru & En$\rightarrow$Ch \\ \midrule 
mBART               & 0.312             & 0.387             & 0.212             & 0.274             & 0.234             & 0.245             & 0.409             \\
mBART+\texttt{graphmax}  & \textbf{0.334}    & \textbf{0.396}    & \textbf{0.219}    & \textbf{0.290}    & \textbf{0.257}    & \textbf{0.258}    & \textbf{0.421}    \\ \midrule
& Cz$\rightarrow$En & Ge$\rightarrow$En & Ha$\rightarrow$En & Ic$\rightarrow$En & Ja$\rightarrow$En & Ru$\rightarrow$En & Ch$\rightarrow$En \\ 
\cmidrule(l){2-8} 
mBART               & 0.276             & 0.361             & 0.252             & 0.334             & \textbf{0.222}    & 0.371             & 0.276             \\
mBART+\texttt{graphmax} & \textbf{0.283}    & \textbf{0.372}    & \textbf{0.273}    & \textbf{0.345}    & 0.220             & \textbf{0.378}    & \textbf{0.299}    \\ \bottomrule
\end{tabular}
\caption{The BLEU scores of mBART and mBART+\texttt{graphmax} on multilingual machine translation.}
\label{tab:machinetransMultipleLan}
\end{table}

\subsection{The $n$-gram Mixture Model}
Section \ref{sec:graphsoftmax} (e.g., Equation (\ref{eq:softmaxGTV})) and Section \ref{sec:experiements} elaborate on that \texttt{graphmax} with a first-order concurrent graph $\tilde{\mathbf{A}}$ acts as a 2-gram model. We can extend it to a general $n$-gram model by replacing $\tilde{\mathbf{A}}$ with a high-order adjacent matrix $\tilde{\mathbf{A}}^{{n}}$. Moreover, instead of using only $\tilde{\mathbf{A}}$, we can employ a mixture model consisting of $\tilde{\mathbf{A}}+\tilde{\mathbf{A}}^2+\cdots+\tilde{\mathbf{A}}^n$. This mixed adjacency matrix integrates more useful knowledge into \texttt{graphmax}. 
The results in Table \ref{tab:2gramVSngram} indicate that an $n$-gram mixture model achieves better performance in most of the experiments involving text generation and machine translation. This further corroborates the effectiveness of incorporating higher-order dependencies in the form of $\tilde{\mathbf{A}}+\tilde{\mathbf{A}}^2$ with \texttt{graphmax}.

\begin{table}[]
    \centering
    \setlength{\tabcolsep}{5mm}{
    \begin{tabular}{llcc}
    \toprule
   Tasks & Metrics & \{2,3\}-gram & \{2,3,4,5\}-gram\\
    \midrule
    & BLEU-2 & 0.989 & 0.992 \\
    & BLEU-3 & 0.945 & 0.965 \\
Text generation      & BLEU-4 & 0.832 & 0.875 \\
  & BLEU-5 & 0.660 & 0.652 \\
    & Perplexity & 84.5 & 85.3\\
    \hline
    Machine translation & BLEU & 0.372 & 0.379\\
    \bottomrule
    \end{tabular} 
    }
    \caption{Comparisons of the $\{2,3\}$-gram model (with $\tilde{\mathbf{A}}$) and the $\{2,3,4,5\}$-gram mixture model (with $\tilde{\mathbf{A}}+\tilde{\mathbf{A}}^2$) on review generation with Amazon and machine translation on WMT'16.}
    \label{tab:2gramVSngram}
\end{table}

\subsection{Human Evaluation}
To evaluate the fluency of the generated text, we further conduct a human evaluation based on the work of \citeA{ahn2016improving}. Participants are presented with a set of food and restaurant reviews generated by GPT-2 (baseline) and our proposed model (GPT-2+\texttt{graphmax}). They are asked to rate the quality of the text based on four perspectives: (1) text style, e.g., a human participant is asked whether the food review generated by the machine is in the social media slang style or in a formal style; (2) {word choice}; (3) {word order variation}; (4) {story-line}, which is a complete narrative logic of the generated text.
In particular, participants read the reviews randomly selected from the baseline and the proposed model, and then rate the text from the four perspectives on a 5-point Likert scale: from 1 (worst) to 5 (best). 

We randomly select 50 food and restaurant reviews generated by GPT-2 and another 50 reviews by the proposed model. The 100 reviews are shuffled and five participants (students) are invited to read and evaluate them independently.
Moreover, we distribute the evaluation task to a wider range of participants through a questionnaire website\footnote{https://www.wjx.cn/}. A total of 27 participants complete the evaluation, including teachers, older adults, and secondary school students. The 100 mixed reviews are shuffled and presented for rating to all 32 participants, resulting in 426 valid ratings.
The overall rating results for the four evaluation categories, as well as their averages, are shown in Table \ref{tab:human_eval}. Our proposed model (GPT-2+\texttt{graphmax}) performs better in controlling the text style and story-line coherence compared with the baseline. Overall, our method generates reviews that are more in line with the social media style and have a more coherent narrative logic.

\begin{table}[]
    \centering
    \begin{tabular}{lcc}
        \toprule
        & \multicolumn{2}{c}{Methods} \\
        \cmidrule{2-3}
        Evaluation Category & GPT-2  & GPT-2+\texttt{graphmax} \\
        \midrule
        Text style & 3.78 (2.56) &  3.94 (1.39) \\
        Word choice & 4.43 (1.89) & 4.31 (1.43)\\
        Word order variation & 4.12 (2.71) & 4.67 (1.67) \\
        Story-line & 2.30 (2.45) & 3.23 (1.41) \\
        \midrule
        Average  & 3.66 & 4.04 \\
        \bottomrule
    \end{tabular}
    \caption{Results of human evaluation on text generation (standard deviations in brackets).}
    \label{tab:human_eval}
\end{table}

\section{Conclusion}
Motivated by
the word concurrent relationships which describe the human preferences of expression, we construct a weighted directed word concurrent graph for scene-specific text generation. The nodes of the graph are the words that appear in the corpus, the direction of edges is determined by the 2-grams of the corpus, and the corresponding weights can be calculated based on the frequency of the 2-grams. 
To incorporate the concurrent information into the process of text generation, we propose a concurrent graph regularized \texttt{softmax} function, called \texttt{graphmax}, for the next word prediction. The proposed \texttt{graphmax} is an $n$-gram mixture model that can be readily plugged into a large-scale pre-trained LM. We apply \texttt{graphmax} to scene-specific review generation and machine translation, and the results from both tasks demonstrate that our method can improve the performances than existing methods using \texttt{softmax}. This suggests that due to its simplicity and versatility \texttt{graphmax} should be used broadly in LMs.  

\section*{Acknowledgements}
We would like to thank the Associate Editor and three referees for their constructive and insightful comments that have significantly improved the paper. 
The research of Guosheng Yin was partially supported by the Theme-based Research Scheme (TRS) from the Research Grants Council of Hong Kong, Institute of Medical Intelligence and XR (T45-401/22-N).

\appendix
\section{Proofs}
\subsection{Equation (\ref{eq:softmax}) in \texttt{Softmax}}
\label{sec:appendix_A}
\begin{proof}
	The augmented Lagrangian function of the optimization problem in Equation (\ref{eq:softmaxOpt}) is
	\begin{equation*}
		L(\mathbf{x}, \lambda) = -\langle \mathbf{x},\mathbf{z}\rangle + \langle \mathbf{x},\log \mathbf{x}\rangle + \lambda (\mathbf{1}^{\top}\mathbf{x} - 1).
	\end{equation*}
	The first-order conditions yield
	\begin{align}
		&\frac{\partial L}{\partial \mathbf{x}}= -\mathbf{z} + \log \mathbf{x} + (1+\lambda)\mathbf{1} = 0,\label{eq:lag1}\\ 
		&\frac{\partial L}{\partial \lambda}= \sum_{i=1}^N x_i-1=0. \label{eq:lag2}
	\end{align}
	
	According to Equation (\ref{eq:lag1}), we have 
	\begin{equation*}\label{eq:4}
		\mathbf{x} = e^{\mathbf{z}-(1+\lambda)\mathbf{1}},
	\end{equation*} 
and its $i$-th component is
	\begin{equation}\label{eq:x_i}
		x_i = e^{z_i}\cdot e^{-(1+\lambda)}
	\end{equation}
	
Plugging Equation (\ref{eq:x_i}) into Equation (\ref{eq:lag2}), we have
	\begin{equation*}
		\sum_{i=1}^N x_i =  e^{-(1+\lambda)} \sum_{i=1}^N e^{z_i} = 1,
	\end{equation*}
	and thus $\sum_{i=1}^N e^{z_i} = e^{(1+\lambda)}$.
By substituting it to Equation (\ref{eq:x_i}), we have $x_i = \frac{e^{z_i}}{\sum_{i=1}^N e^{z_i}}$, which completes the proof.
\end{proof}

\subsection{Proof of Proposition 2}
\label{sec:appendix_B}
\begin{proof}
    For $a_i < a_j$, the first inequality is $\texttt{graphmax}_j(\mathbf{a}) - \texttt{graphmax}_i(\mathbf{a})  \geq 0 $. A proof of contradiction can be easily constructed. Suppose that for $a_i < a_j$ we have $\texttt{graphmax}_j(\mathbf{a}) - \texttt{graphmax}_i(\mathbf{a}) < 0 $.
From the definition of the projection operator $\Pi_{\Delta}(\mathbf{a})=\argmin_{\mathbf{x}\in \Delta} \|\mathbf{x}-\mathbf{a}\|_2$ in Section \ref{sec:projectionDelta},  we have
\begin{equation*}
    \|\mathbf{x}-\mathbf{a}\|_2\\ \geq \|\texttt{graphmax}(\mathbf{a})-\mathbf{a}\|_2\\,
\end{equation*}
for $\mathbf{x}\in \Delta^{N-1}$.
This leads to a contradiction,
\begin{align*}
    \mathbf{a} &= (a_1,a_2,\ldots,a_i,\ldots,a_j,\ldots)\\
    \texttt{graphmax}(\mathbf{a}) & = (a_1,a_2,\ldots,a_j,\ldots,a_i,\ldots).
\end{align*}
Concretely, 
\begin{align*}
    &\|\mathbf{x}-\mathbf{a}\|_2 - \|\texttt{graphmax}(\mathbf{a})-\mathbf{a}\|_2\\
    =& \sum_{i=1}^N (x_i-a_i)^2 - \sum_{i=1}^N(\texttt{graphmax}_i(\mathbf{a})-a_i)^2\\
     =& (\texttt{graphmax}_j(\mathbf{a}) - a_i)^2 + (\texttt{graphmax}_i(\mathbf{a}) - a_j)^2 \\
     &-(\texttt{graphmax}_i(\mathbf{a}) - a_i)^2 - (\texttt{graphmax}_j(\mathbf{a}) - a_j)^2 \\
     =& 2(\texttt{graphmax}_i(\mathbf{a}) - \texttt{graphmax}_j(\mathbf{a})) (a_i - a_j) < 0
\end{align*}
where $\texttt{graphmax}_i(\mathbf{a})$ denotes the $i$-th component of the vector $\texttt{graphmax}(\mathbf{a})$. 

The second inequality $\texttt{graphmax}_j(\mathbf{a}) - \texttt{graphmax}_i(\mathbf{a})  \leq a_j -  a_i$ contains three cases that can be proved one by one as follows.

For case 1, 
\begin{equation*}
	\begin{cases}
		a_i + \gamma \geq 0,\\
		a_j + \gamma \geq 0,
	\end{cases}
\end{equation*}
according to the definition of $\texttt{graphmax}$ in Equation (\ref{eq:graphsoftDefinition}), we have 
\begin{equation*}
    \texttt{graphmax}_j(\mathbf{a}) - \texttt{graphmax}_i(\mathbf{a}) = a_j - a_i <0.
\end{equation*}

For case 2,
\begin{equation*}
	\begin{cases}
		a_i + \gamma \leq 0,\\
		a_j + \gamma \geq 0,
	\end{cases}
\end{equation*}
we have 
\begin{equation*}
    \texttt{graphmax}_j(\mathbf{a}) - \texttt{graphmax}_i(\mathbf{a}) = a_j + \gamma \leq a_j-a_i <0 .
\end{equation*}

Finally, for case 3, we have
\begin{equation*}
	\begin{cases}
		a_i + \gamma < 0,\\
		a_j + \gamma < 0,
	\end{cases}
\end{equation*}
and this leads to
\begin{equation*}
    \texttt{graphmax}_j(\mathbf{a}) - \texttt{graphmax}_i(\mathbf{a}) = 0< a_j-a_i.
\end{equation*}
This completes the proof.
\end{proof}

\vskip 0.2in
\bibliography{sample}
\bibliographystyle{theapa}

\end{document}